%% file: main.tex
\documentclass[11pt]{article}

\usepackage[preprint]{acl}

\usepackage{times}
\usepackage{latexsym}

\usepackage[T1]{fontenc}

\usepackage[utf8]{inputenc}
\usepackage{subcaption}
\usepackage{booktabs} 

\usepackage{microtype}
\usepackage{xcolor}
\usepackage[table]{xcolor}
\usepackage{marvosym}
\usepackage{multirow}

\usepackage{inconsolata}

\usepackage{graphicx}
\usepackage{amsmath,amssymb}

\definecolor{myred}{HTML}{C44E52}
\definecolor{mygreen}{HTML}{55A868}

\title{Pair-In, Pair-Out: Latent Multi-Token Prediction for Efficient LLMs}

\author{
 \textbf{Wenhui Tan\textsuperscript{1}},
 \textbf{Minghao Li\textsuperscript{2}},
 \textbf{Xiaoqian Ma\textsuperscript{2}},
 \textbf{Siqi Fan\textsuperscript{3}},
\\
 \textbf{Xiusheng Huang\textsuperscript{4}},
 \textbf{Liujie Zhang\textsuperscript{2}},
 \textbf{Ruihua Song\textsuperscript{1}},
 \textbf{Weihang Chen\textsuperscript{2}}
\\
 \textsuperscript{1}Gaoling School of Artificial Intelligence, Renmin University of China,\\
 \textsuperscript{2}AI Platform, Xiaohongshu Inc.,\\
 \textsuperscript{3}University of Electronic Science and Technology of China,\\
 \textsuperscript{4}Institute of Automation, Chinese Academy of Sciences
\\
 \small{
   \textbf{Correspondence:} \href{mailto:rsong@ruc.edu.cn}{Ruihua Song} and \href{mailto:chenjinzhi@xiaohongshu.com}{Weihang Chen}
 }
 \\
 \small{
   \textbf{Project Page:} \href{https://github.com/RedAI-Infra/PIPO}{GitHub.com/RedAI-Infra/PIPO}
 }
}

\begin{document}
\maketitle

\begin{abstract}
Long chain-of-thought reasoning has made autoregressive decoding the dominant inference cost of modern large language models.
Existing methods target either the input side (latent compression) or the output side (speculative decoding and multi-token prediction, MTP), but the two lines of work have been pursued independently.
Moreover, output-side methods must incur an expensive verifier pass to validate the unreliable draft tokens predicted by MTP.
To address these issues, we propose \textbf{Pair-In, Pair-Out (PIPO)}, which unifies both sides by viewing a latent compressor and an MTP head as mirror-image operations: the compressor folds two input tokens into one latent representation, while the MTP head unfolds one hidden state into one additional output token.
To remove the verifier cost without sacrificing reliability, PIPO trains a lightweight confidence head that decides whether draft tokens should be accepted.
We observe that On-Policy Distillation (OPD) naturally matches the rejection-sampling criterion of speculative decoding, so the confidence head can be trained alongside OPD with negligible extra cost.
Experiments on AIME 2025, GPQA-Diamond, LiveCodeBench v6, and LongBench v2 with Qwen3.5-4B and 9B backbones show that PIPO improves pass@4 over regular decoding by up to $+7.15$ points, while delivering up to $2.64\times$ first-token-latency and $2.07\times$ per-token-latency speedups.
\end{abstract}

\input{sections/1-introduction.tex}
\input{sections/2-related_work.tex}
\input{sections/3-method.tex}
\input{sections/4-experiments.tex}
\input{sections/5-conclusion.tex}

\input{sections/6-limitation_ethics.tex}

\bibliography{custom}
\newpage
\appendix

\input{sections/9-appendix}

\end{document}

%% file: sections/1-introduction.tex
\begin{figure*}[t]
\centering
\includegraphics[width=\textwidth]{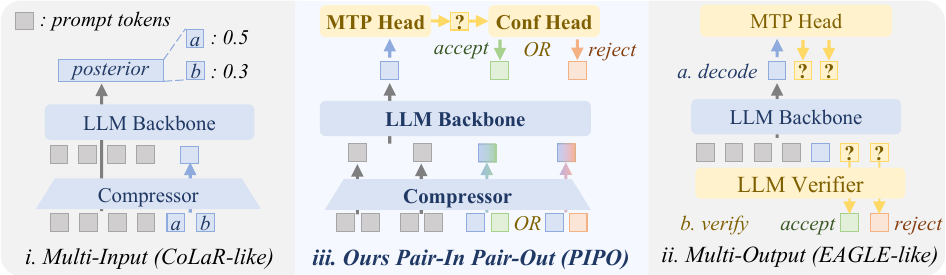}
\caption{Comparison of input-side methods, output-side methods, and our proposed PIPO.
PIPO treats a latent compressor and a multi-token prediction (MTP) head as mirror-image operations around the backbone, and trains a lightweight confidence head that replaces the verifier of speculative decoding (e.g., EAGLE).}
\label{fig:teaser}
\end{figure*}

\section{Introduction}\label{sec:introduction}
Large language models (LLMs) are increasingly used as reasoners on complex tasks such as mathematics and coding, typically by generating long chains of intermediate reasoning tokens before producing a final answer~\cite{cot,o1,r1}.
While this paradigm improves accuracy, it makes inference expensive.
Under standard autoregressive decoding, an LLM emits one token per forward step, and each step depends on all previously generated tokens.
Long reasoning traces therefore translate directly into long decoding latency, and the per-token cost has become the dominant inference bottleneck of modern reasoning LLMs.

From a model-architecture perspective, existing methods address this bottleneck from two sides.
\textbf{Output-side} methods predict more tokens per forward step:
speculative decoding~\cite{speculativedecoding} predicts draft tokens with a small draft model, and verifies them with a large verifier model, which determines whether to accept draft tokens via rejection sampling;
EAGLE~\cite{eagle} extends this idea to the hidden-feature level, and verifies a tree of candidates in parallel, yielding higher acceptance rates.
Recent strong backbones incorporate multi-token prediction (MTP) heads directly into the model~\cite{deepseekv3,mimov2flash,qwen35}, which serve as co-trained drafters within the same speculative framework.
\textbf{Input-side} methods reduce the effective sequence length with a compressor, merging multiple input tokens into one latent / continuous representation~\cite{colar,softthinking,multiplexthinking}, i.e., latent reasoning~\cite{latentspacesurvey}.

These two lines, however, have been pursued independently, and a unified design that exploits \emph{both} sides is still missing.
Moreover, output-side speedups remain bounded by the verifier's forward-pass cost: every accepted token must still pass through the full backbone.
To bridge these gaps, we start from two key observations.

\begin{figure}[t]
\centering
\includegraphics[width=0.8\linewidth]{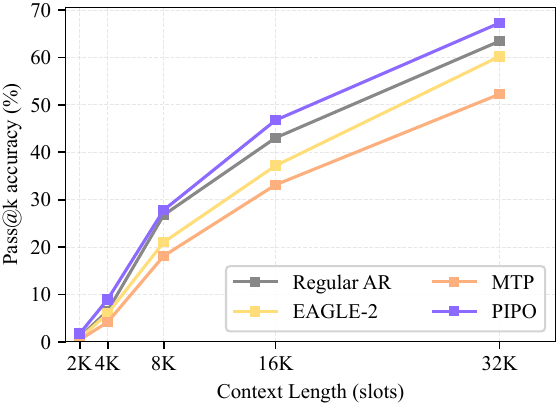}
\caption{Overall pass@4 of PIPO and baseline methods on the four evaluation benchmarks, as the response context budget varies from $2$K to $32$K.}
\label{fig:acc_length}
\end{figure}

\paragraph{Observation 1: a latent compressor and an MTP head are mirror images.}
A latent compressor folds two input token embeddings into one latent representation on the way in; an MTP head unfolds one hidden state into one additional output token on the way out.
Combining them yields a symmetric \textbf{pair-in / pair-out} interface that simultaneously halves the effective input length and doubles the per-step output.
We call this framework \textbf{Pair-In, Pair-Out (PIPO)} (Figure~\ref{fig:teaser}).
Because MTP heads are already included in modern strong LLMs, PIPO can be built from mostly off-the-shelf components.

\paragraph{Observation 2: the on-policy distillation teacher \emph{is} the speculative-decoding verifier.}
In \textbf{speculative decoding}, a strong LLM serves as a \textbf{verifier} that accepts a draft token $x$ with probability $\min(p_v(x)/p_d(x),\,1)$, where $p_v(x)$ is the verifier's probability of $x$, and $p_d(x)$ is the draft model's probability of $x$.
In \textbf{on-policy distillation (OPD)}~\cite{opd,rethinkingopd}, a strong model serves as a \textbf{teacher} to drive a reverse-KL distillation loss $\mathcal{D}\!=\!\mathrm{KL}(p_s\,\Vert\,p_t)$, where $p_t$ is the teacher's probability and $p_s$ is the student's probability.
These two roles play the same function: both judge whether the student / draft is consistent with a stronger teacher / verifier.
PIPO exploits this by integrating a lightweight \textbf{confidence head}, trained with the rejection-sampling acceptance probability $\min(p_t/p_s,\,1)$ as its label.
The supervision is essentially free: the same $(p_t, p_s)$ distributions are already computed for OPD training.
At inference time, the lightweight confidence head replaces the heavy verifier pass.

We evaluate PIPO on AIME 2025~\cite{aime}, GPQA-Diamond~\cite{gpqa}, LiveCodeBench v6~\cite{livecodebench}, and LongBench v2~\cite{longbenchv2}, using Qwen3.5-4B and Qwen3.5-9B backbones~\cite{qwen35}, against regular decoding, MTP decoding without verification, and EAGLE-2~\cite{eagle2} speculative decoding.
PIPO is the strongest pass@4 method on both backbones, improving the best baseline by $\mathbf{+3.83}$ points on Qwen3.5-4B and $\mathbf{+7.15}$ points on Qwen3.5-9B.
As shown in Figure~\ref{fig:acc_length}, this advantage \emph{widens} as the context budget grows, because PIPO emits twice as many tokens per step, thus fits more complete chains of reasoning into the same budget.
On the efficiency side, PIPO achieves up to a $2.64\times$ speedup in time-to-first-token (TTFT) and a $2.07\times$ speedup in time-per-output-token (TPOT) over regular decoding.

To conclude, our contributions are:
\begin{itemize}
    \item We propose PIPO, which unifies input-side latent compression with output-side multi-token prediction for efficient LLM decoding.
    \item We observe that the OPD teacher and the speculative-decoding verifier play the same role, and exploit this to train a lightweight confidence head.
    \item Across four challenging benchmarks on Qwen3.5-4B and 9B, PIPO improves pass@4 by up to $+7.15$ points over regular decoding, while delivering up to $2.64\times$ TTFT and $2.07\times$ TPOT speedups.
\end{itemize}

%% file: sections/2-related_work.tex
\section{Related Work}\label{sec:related_work}
\subsection{Output-side: Multi-token Decoding}
Output-side accelerators aim to emit more than one token per forward step.
\emph{Speculative decoding} drafts candidate tokens with a smaller proposal model and verifies them by running a full forward pass through the target backbone~\cite{speculativedecoding}.
EAGLE and its variants improve draft quality by drafting in feature space and verifying dynamic draft trees in a single parallel pass~\cite{eagle,eagle2,eagle3}.
These methods preserve the target distribution exactly, but the verification pass through the large backbone is incurred at \emph{every} decoding step, and at long context lengths this verifier cost can erase the gain from accepting multiple tokens at once.

A parallel line attaches additional prediction heads directly to the backbone.
Medusa adds multiple decoding heads for parallel token prediction~\cite{medusa}, and recent large models go further by natively including MTP modules during pretraining or post-training~\cite{deepseekv3,qwen3,mimov2flash}, turning the MTP head into a standard, built-in component of modern LLMs.
These heads reduce the number of decoding steps when their drafts are accepted, but without a verification or acceptance mechanism, unreliable drafts propagate into the generated sequence and hurt downstream accuracy.

PIPO builds on the strength of pretrained MTP heads and resolves the verifier bottleneck differently.
Instead of an extra backbone forward pass per step, PIPO learns a lightweight \emph{confidence head} whose supervision is recycled, for free, from the teacher distributions of on-policy distillation (see Section~\ref{sec:method:opd}).
The trained head replaces the per-step verifier entirely, turning a recurring inference cost into a one-time training signal.

\subsection{Input-side: Latent Compression}
Input-side methods reduce the number of effective input tokens by replacing or compressing token-level reasoning with continuous representations.
Coconut feeds the backbone's hidden states back to the model as continuous thoughts in place of decoded tokens~\cite{coconut}.
Soft Thinking and related analyses study soft or stochastic token representations in continuous concept space~\cite{softthinking,singlethreaded,multiplexthinking}.
These methods change \emph{what} the model attends to, but do not change \emph{how many} tokens are emitted per decoding step, and so leave the output-side bottleneck untouched.

CoLaR~\cite{colar} is the closest work to PIPO: it compresses reasoning chains in latent space and directly predicts compressed latent embeddings.
However, its latent prediction relies on a unimodal Gaussian assumption, which poorly fits the multi-modal distribution of multi-token continuations.
PIPO keeps generation in the discrete vocabulary space, and uses latent variables only on the \emph{input} side.
This asymmetric design is what makes pair-in compression compatible with an off-the-shelf MTP head and a confidence-based acceptance rule.
Additional related work on reasoning and reasoning-efficient LLMs is discussed in Appendix~\ref{app:sec:relatedwork}.

%% file: sections/3-method.tex
\section{Method}\label{sec:method}

\begin{figure*}
    \centering
    \includegraphics[width=\linewidth]{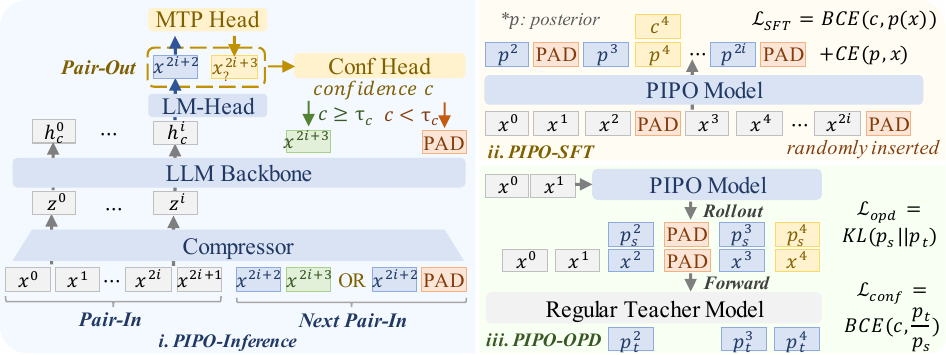}
    \caption{Illustration of PIPO's architecture (i), SFT training process (ii), and OPD training process (iii).}
    \label{fig:method}
\end{figure*}

\subsection{Overview and Notation}
PIPO changes the decoding unit from a single token to a token \emph{pair}.
Let $x^j$ denote the embedding at position $j$; we use the same symbol for a \emph{token} and its \emph{embedding} when the meaning is clear, since the embedding layer is fixed.

As illustrated in Figure~\ref{fig:method}\,(i), at pair step $i$ PIPO receives a compressed latent input $z^i$ that represents two consecutive tokens $(x^{2i}, x^{2i+1})$, and predicts a backbone-token distribution $p_b^{2i+2}$ together with a draft-token distribution $p_d^{2i+3}$.
To gain efficiency without committing to unreliable drafts, PIPO also predicts a confidence score $c^i \in (0,1)$.
If $c^i \geq \tau_c$, the draft token is accepted and the next input pair contains two newly generated tokens.
Otherwise, the draft token is rejected and replaced by a padding token before compression, so that PIPO keeps the same pair-level interface regardless of acceptance.

\subsection{The Pair-In / Pair-Out Architecture}\label{sec:method:pipo}
The PIPO architecture is built from two operations that are symmetric around the backbone: a compressor on the input side and an MTP head on the output side.

\paragraph{Pair-in compression.}
The compressor maps every two consecutive token embeddings into one latent input:
\begin{equation}
    z^i = f_{\theta}\!\left([x^{2i}; x^{2i+1}]\right),
\end{equation}
where $[\cdot\,;\cdot]$ denotes concatenation and $f_{\theta}$ is an MLP.
We initialize $f_{\theta}$ so that $f_{\theta}([a;b]) \approx a + b$, which keeps the backbone's input distribution close to its pretraining distribution at the start of training (discussed in Section~\ref{sec:experiments:ablation}).
Appendix~\ref{app:sec:experiments:compressor} probes the \emph{trained} compressor and shows it preserves the \emph{additive geometry} while learning a \emph{non-symmetric}, \emph{position-aware} projection.

\paragraph{Pair-out prediction.}
The LLM backbone produces a hidden state over the compressed prefix and a distribution over the next backbone token:
\begin{align}
    h_b^i &= \mathrm{Backbone}(z^{\leq i}), \\
    p_b^{2i+2} &= \mathrm{LMHead}(h_b^i).
\end{align}
Then, the draft token is predicted by an MTP head, conditioned on the backbone hidden state and the embedding of the backbone token:
\begin{align}
    h_d^i &= \mathrm{MTPHead}(h_b^i, x^{2i+2}), \\
    p_d^{2i+3} &= \mathrm{LMHead}(h_d^i).
\end{align}

Compression and MTP prediction are mirror-image operations: one folds two token embeddings into one latent input, and the other unfolds one hidden state into one additional output token.
MTP heads are already included in modern strong LLMs, so the only new module that PIPO adds to the backbone is the small MLP compressor and confidence head (as discussed below).
PIPO therefore obtains a pair-in / pair-out interface mostly from off-the-shelf components.

\paragraph{Confidence-guided draft acceptance.}
In speculative decoding, the draft token can be wrong and must be verified by a large verifier model, which is computationally expensive.
PIPO instead integrates a lightweight confidence head that directly estimates whether the draft token should be used:
\begin{equation}
    c^i = g_{\phi}\!\left([h_b^i; h_d^i]\right),
\end{equation}
where $g_\phi$ is a small MLP that produces a scalar in $(0,1)$.
If $c^i \geq \tau_c$, the next input pair is $(x^{2i+2}, x^{2i+3})$; otherwise it is $(x^{2i+2}, x^{\mathrm{pad}})$, where $x^{\mathrm{pad}}$ is a fixed padding embedding.
This rule provides a safe fallback to regular single-token decoding whenever the head is uncertain, while incurring only one extra MLP forward pass per pair instead of an entire backbone pass.
Appendix~\ref{app:sec:experiments:pad} verifies that the compressor effectively treats a padded pair as the surviving token alone, so the fallback does not pollute the backbone's hidden state.

\subsection{Supervised Fine-Tuning}\label{sec:method:sft}
SFT trains PIPO with a next-pair prediction objective.
At each pair step $i$, the model predicts the ground-truth backbone token $x^{2i+2}$ and draft token $x^{2i+3}$ with a standard cross-entropy (CE) loss:
\begin{equation}
    \mathcal{L}_{\mathrm{tok}} =
    \mathrm{CE}(p_b^{2i+2}, x^{2i+2}) +
    \mathrm{CE}(p_d^{2i+3}, x^{2i+3}).
\end{equation}
We additionally bootstrap the confidence head in this stage: its prediction $c^i$ is aligned with the ground-truth draft token's probability through a binary cross-entropy (BCE) loss,
\begin{equation}
    \mathcal{L}_{\mathrm{conf}}^{\mathrm{SFT}} =
    \mathrm{BCE}\!\left(c^i,\, p_d^{2i+3}(x^{2i+3})\right),
\end{equation}
so that $c^i$ already correlates with draft reliability before OPD provides a sharper supervision signal.
The overall SFT objective is $\mathcal{L}_{\mathrm{SFT}} = \mathcal{L}_{\mathrm{tok}} + \lambda_{\mathrm{conf}}\, \mathcal{L}_{\mathrm{conf}}^{\mathrm{SFT}}$.
To match inference behavior, we randomly replace a fraction of the draft-position inputs with the padding embedding $x^{\mathrm{pad}}$, exposing the model to rejected-draft cases during training.

\subsection{On-Policy Distillation}\label{sec:method:opd}
SFT trains the pair-level interface, but never exposes PIPO to its own decoding distribution.
We close this train--inference gap with on-policy distillation (OPD).
Given a prompt, PIPO rolls out a response under its current policy, recording the accepted draft tokens, the padding tokens at rejected draft positions, and the student distributions $p_s$ and confidence scores $c$ at every step.
We then remove the padding tokens, feed the resulting clean text to an uncompressed teacher, and obtain token-level teacher distributions $p_t$.
The reverse-KL distillation loss aligns the student to the teacher,
\begin{equation}
    \mathcal{L}_{\mathrm{distill}} = \mathrm{KL}(p_s \,\Vert\, p_t).
\end{equation}

\paragraph{The teacher is the verifier.}
Speculative decoding accepts a draft token $x$ with probability $\min(p_t(x)/p_s(x), 1)$, using a strong target model as the per-step verifier.
In OPD, the same $(p_t, p_s)$ pair is already computed at every position, i.e., exactly the quantities required by the verifier criterion.
The OPD teacher and the speculative-decoding verifier therefore play the same role: both judge whether the student draft is consistent with a stronger reference distribution.

PIPO exploits this role-unification by training the confidence head with the rejection-sampling acceptance probability:
\begin{align}
    y_c &= \min\left(\frac{p_t(x)}{p_s(x)}, 1\right), \\
    \mathcal{L}_{\mathrm{conf}}^{\mathrm{OPD}} &= \mathrm{BCE}(c, y_c).
\end{align}
This supervision is essentially free: $(p_t, p_s)$ are already computed for $\mathcal{L}_{\mathrm{distill}}$, so the confidence head reuses them with no extra forward passes and no extra labels.
At inference, the trained confidence head replaces the verifier pass entirely, turning a per-step inference cost into a one-time training signal.
Overall, the OPD objective is
\begin{equation}
    \mathcal{L}_{\mathrm{OPD}} = \mathcal{L}_{\mathrm{distill}} + \lambda_{\mathrm{conf}} \mathcal{L}_{\mathrm{conf}}^{\mathrm{OPD}}.
\end{equation}

%% file: sections/4-experiments.tex
\section{Experiments}\label{sec:experiments}

\begin{table*}[ht]
\small
\centering
\caption{Main results on AIME 2025, GPQA-Diamond, LiveCodeBench v6, and LongBench v2 (short).
All methods share the same Qwen3.5-4B / 9B backbone and a $32$K-slot response budget.
\textbf{Bold} marks the best score in each column; \underline{underline} marks the second best.}
\label{tab:main}
\begin{tabular}{lcccccccccc}
\toprule
\multicolumn{1}{l|}{}         & \multicolumn{2}{c|}{AIME 2025}      & \multicolumn{2}{c|}{GPQA-Diamond}   & \multicolumn{2}{c|}{LiveCodeBench v6} & \multicolumn{2}{c|}{LongBench v2}   & \multicolumn{2}{c}{Overall} \\
\multicolumn{1}{l|}{}         & avg@4 & \multicolumn{1}{c|}{pass@4} & avg@4 & \multicolumn{1}{c|}{pass@4} & avg@4  & \multicolumn{1}{c|}{pass@4}  & avg@4 & \multicolumn{1}{c|}{pass@4} & avg@4        & pass@4       \\ \midrule
\multicolumn{11}{l}{\textit{\textbf{Qwen3.5-4B}}}                                                                                                                                                                     \\ \midrule
\multicolumn{1}{l|}{Regular}  & \underline{49.17} & \multicolumn{1}{c|}{\underline{63.33}}  & \textbf{59.72} & \multicolumn{1}{c|}{\underline{72.73}}  & \textbf{40.08}  & \multicolumn{1}{c|}{44.27}   & \textbf{59.69} & \multicolumn{1}{c|}{\textbf{73.60}}  & \textbf{52.16}        & 63.48        \\
\multicolumn{1}{l|}{Eagle-2}  & 40.83 & \multicolumn{1}{c|}{60.00}  & 53.66 & \multicolumn{1}{c|}{66.16}  & \underline{32.06}  & \multicolumn{1}{c|}{43.51}   & \underline{58.29} & \multicolumn{1}{c|}{71.35}  & 46.21        & 60.26        \\
\multicolumn{1}{l|}{MTP}      & 34.17 & \multicolumn{1}{c|}{43.33}  & 52.27 & \multicolumn{1}{c|}{69.19}  & 14.89  & \multicolumn{1}{c|}{27.48}   & 52.67 & \multicolumn{1}{c|}{69.10}  & 38.50        & 52.28        \\
\rowcolor{blue!5}
\multicolumn{1}{l|}{\textbf{PIPO-SFT}} & 42.50 & \multicolumn{1}{c|}{60.00}  & \underline{59.47} & \multicolumn{1}{c|}{\textbf{79.29}}  & 30.15  & \multicolumn{1}{c|}{\underline{48.85}}   & 49.02 & \multicolumn{1}{c|}{\underline{71.91}}  & 45.28 & \underline{65.01} \\
\rowcolor{blue!10}
\multicolumn{1}{l|}{  \textbf{+ OPD}} & \textbf{50.00} & \multicolumn{1}{c|}{\textbf{76.67}}     & 54.17 & \multicolumn{1}{c|}{\underline{72.73}}  & \underline{32.06} & \multicolumn{1}{c|}{\textbf{49.62}}    & 49.86 & \multicolumn{1}{c|}{70.22}       & \underline{46.52}  & \textbf{67.31} \\ \midrule
\multicolumn{11}{l}{\textit{\textbf{Qwen3.5-9B}}}                                                                                                                                                                     \\ \midrule
\multicolumn{1}{l|}{Regular}  & \underline{53.33} & \multicolumn{1}{c|}{66.67}  & \textbf{68.56} & \multicolumn{1}{c|}{78.79}  & \textbf{47.14}  & \multicolumn{1}{c|}{54.20}    & \textbf{61.66} & \multicolumn{1}{c|}{72.47}  & \textbf{57.67}        & 68.03        \\
\multicolumn{1}{l|}{Eagle-2}  & 49.17 & \multicolumn{1}{c|}{63.33}  & 62.63 & \multicolumn{1}{c|}{73.23}  & 39.31  & \multicolumn{1}{c|}{45.80}   & \underline{61.24} & \multicolumn{1}{c|}{71.35}  & 53.09        & 63.43        \\
\multicolumn{1}{l|}{MTP}      & 40.00 & \multicolumn{1}{c|}{50.00}  & 55.30  & \multicolumn{1}{c|}{71.21}  & 18.32  & \multicolumn{1}{c|}{31.30}   & 52.81 & \multicolumn{1}{c|}{71.91}  & 41.61        & 56.10        \\
\rowcolor{blue!5}
\multicolumn{1}{l|}{\textbf{PIPO-SFT}} & 51.67  & \multicolumn{1}{c|}{\underline{76.67}} & 63.76 & \multicolumn{1}{c|}{\underline{81.31}}  & 34.92  & \multicolumn{1}{c|}{\underline{54.20}}    & 54.35 & \multicolumn{1}{c|}{\textbf{74.16}}  & 51.18  & \underline{71.58} \\
\rowcolor{blue!10}
\multicolumn{1}{l|}{  \textbf{+ OPD}} &  \textbf{59.17} & \multicolumn{1}{c|}{\textbf{83.33}}  & \underline{67.17} & \multicolumn{1}{c|}{\textbf{82.32}}  & \underline{46.56} & \multicolumn{1}{c|}{\textbf{62.60}}     & 56.46  & \multicolumn{1}{c|}{\underline{72.47}}  & \underline{57.34} & \textbf{75.18} \\ \bottomrule
\end{tabular}
\end{table*}

\subsection{Experimental Setup}

\paragraph{Training data.}
We train PIPO on DAPO-Math~\cite{dapo} ($17.4$k math questions) and Codeforces~\cite{codeforces} ($16.1$k coding questions), with a $90$\,/\,$10$ SFT\,/\,OPD split.
SFT trajectories come from sampling four responses per question with Qwen3.5-9B (the teacher) and keeping all correct ones, yielding $\sim\!90$k trajectories of average length $24.4$K tokens (capped at $64$K; full statistics of SFT training data are in Appendix~\ref{app:sec:training}).
For OPD we additionally roll out four teacher responses per question to estimate difficulty and to drive the data filter studied in Section~\ref{sec:experiments:ablation}.

\paragraph{Training setting.}
PIPO is trained in two stages: $2$ epochs of SFT with $25$\% random padding at draft positions (to expose the model to rejected-draft inputs), followed by $1$ epoch of OPD on rollouts of the SFT student.
Both stages use LoRA~\cite{lora} adapters and AdamW with learning rate $1\!\times\!10^{-4}$, $5$\% warmup and cosine annealing; the confidence-loss weight is fixed to $\lambda_{\mathrm{conf}}\!=\!1.0$ throughout.
Further details are elaborated in Appendix~\ref{app:sec:training}.

\paragraph{Evaluation data.}
We evaluate on four challenging benchmarks: (i) \emph{AIME 2025}~\cite{aime} consists of $30$ math competition problems;
(ii) \emph{GPQA-Diamond}~\cite{gpqa} consists of $198$ graduate-level multiple-choice questions in physics, chemistry, and biology;
(iii) \emph{LiveCodeBench v6}~\cite{livecodebench} consists of $131$ recent competitive-programming problems; and
(iv) \emph{LongBench v2}~\cite{longbenchv2} (short subset due to the models' context limit) consists of $178$ long-context ($>$10K input) reasoning problems.

\paragraph{Evaluation setting.}
Following Qwen3.5~\cite{qwen35}, we use $\mathrm{temperature}\!=\!1.0$, $\mathrm{top\text{-}p}\!=\!0.95$, $\mathrm{top\text{-}k}\!=\!20$, and a repetition penalty of $1.5$, with a $32$K-slot response budget shared by all methods (the semantics of \emph{slot} is discussed in Section~\ref{sec:experiments:efficiency}).
We sample four responses per question and report \textbf{avg@4} (mean accuracy) and \textbf{pass@4} (at least one of the four trials correct); further details are elaborated in Appendix~\ref{app:sec:evaluation}.

\paragraph{Baselines.}
All methods run on the same Qwen3.5-4B and 9B backbones~\cite{qwen35}.
We compare PIPO against (i) \textbf{Regular} autoregressive decoding; (ii) \textbf{MTP}~\cite{qwen35}, which uses the pretrained MTP head to emit a draft token per step \emph{without verification}, doubling per-step output at the risk of propagating unreliable drafts; and (iii) \textbf{EAGLE-2}~\cite{eagle2}, a strong speculative-decoding baseline that drafts with the MTP head and verifies the draft tree in one backbone forward pass.
We report two PIPO variants: \textbf{PIPO-SFT} (SFT only) and \textbf{PIPO\,+\,OPD} (on-policy distillation post-trained on PIPO-SFT).

\subsection{Main Results}\label{sec:experiments:main}
To evaluate the effectiveness of PIPO, we compare it against three strong baselines on four challenging benchmarks.
Table~\ref{tab:main} reports the main results, from which we draw three key observations.

\paragraph{PIPO is the strongest pass@4 method on both backbones.}
Even without OPD, PIPO-SFT already surpasses every baseline on pass@4, improving over the best baseline (Regular) by $+1.53$ points on Qwen3.5-4B and $+3.55$ points on Qwen3.5-9B.
Adding OPD widens the gain to $+3.83$ points on $4$B and $+7.15$ points on $9$B, with the best pass@4 in every per-task column except $4$B LongBench.
We attribute this to PIPO's pair-in interface: under a fixed $32$K-slot response budget, doubling the per-step output halves the effective length cost of a token, letting PIPO fit more complete reasoning chains into the same budget.
The effect is most visible on AIME 2025, where PIPO\,+\,OPD lifts pass@4 by $+13.34$ and $+16.66$ points on $4$B and $9$B, respectively.

\paragraph{OPD recovers avg@4 while preserving the pass@4 gain.}
PIPO-SFT trades some avg@4 for higher pass@4, because doubling the per-step output adds uncertainty at draft positions.
OPD closes this gap: $+1.24$ avg@4 / $+2.30$ pass@4 on $4$B, and $+6.16$ avg@4 / $+3.60$ pass@4 on $9$B.
On the $9$B backbone, PIPO\,+\,OPD matches Regular on avg@4 ($57.34$ vs.\ $57.67$) while improving pass@4 by $+7.15$ points.
Distillation from the teacher restores stability without sacrificing answer-space coverage.
The SFT-to-OPD gain also grows with model size, suggesting that larger backbones benefit more from OPD.

\paragraph{Existing accelerators still trade off accuracy.}
MTP, which accepts every draft without verification, drops pass@4 by more than $11$ points on both backbones, confirming that unverified drafts propagate errors.
EAGLE-2, with its verifier in the loop, is closer to Regular but still $3$--$5$ points worse on pass@4.\footnote{EAGLE-2's acceptance rule is distribution-preserving only under \emph{exact} (greedy or temperature-only) speculative sampling~\cite{speculativedecoding}; the truncation-based samplers we use (top-$p$, top-$k$, repetition penalty) fall outside this guarantee, so the draft no longer matches the  verifier distribution, and practical acceptance becomes only approximately lossless. Small per-token drifts then accumulate over multi-thousand-token reasoning traces.}
PIPO replaces this verifier with a single MLP per pair, gaining both higher pass@4 and the efficiency profile reported next.

\subsection{Efficiency Analysis}\label{sec:experiments:efficiency}
\begin{figure}[t]
    \centering
    \includegraphics[width=\linewidth]{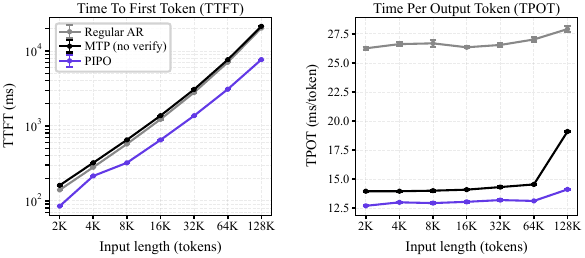}
    \caption{TTFT and TPOT on Qwen3.5-4B at different levels of input length, averaged over $16$ trials.}
    \label{fig:efficiency}
\end{figure}

\begin{table}[t]
\centering
\small
\caption{Average output slots (\# $L$) per response on Qwen3.5-4B/9B under a $32$K-slot generation budget.
Reg.\,=\,Regular, E-2\,=\,EAGLE-2, PIPO-S/-O\,=\,PIPO-SFT / PIPO\,+\,OPD.}
\label{tab:length}
\begin{tabular}{l|ccccc}
\toprule
\# $L$ & Reg. & E-2 & MTP   & PIPO-S & PIPO-O \\ \midrule
4B       & 20{,}160   & 21{,}667   & 22{,}072 & \textbf{18{,}082}    & 19{,}431    \\
9B       & 19{,}140   & 20{,}584   & 21{,}857 & \textbf{17{,}494}    & 18{,}590    \\ \bottomrule
\end{tabular}
\end{table}

We assess efficiency along two complementary axes: output slots per response (Table~\ref{tab:length}) and wall-clock latency on the HuggingFace backend at input lengths $\{2,4,8,16,32,64,128\}$K (Figure~\ref{fig:efficiency}).
For wall-clock we report \textbf{TTFT} (time-to-first-token) and \textbf{TPOT} (time per output token, averaged over the first $16$ generated tokens), each averaged over $16$ trials.

\paragraph{Slot efficiency.}
A \emph{slot} denotes one output unit at the method's native granularity: one decoded token for Regular, EAGLE-2 and MTP (each emitted token occupies one slot); one \emph{token-pair} for PIPO, which emits one backbone token and one draft token per step, which are then compressed back into one input latent for the next step.
The tokens-per-slot ratio is therefore fixed at $1\times$ for baselines, but ranges between $1\times$ and $2\times$ for PIPO depending on the confidence-head acceptance rate.
Under the shared $32$K-slot generation budget, PIPO\,+\,OPD still uses \emph{fewer slots} than baselines: $\sim\!3\%$ fewer than Regular, $\sim\!10\%$ fewer than EAGLE-2, and $\sim\!13\%$ fewer than MTP on both backbones, with PIPO-SFT cutting further to $-10\%$ on $4$B and $-9\%$ on $9$B, while each slot still contributes \emph{strictly more} reasoning content than a baseline slot.
OPD trades a few additional slots for the higher pass@4 reported in Section~\ref{sec:experiments:main}.

\paragraph{TTFT.}
Regular decoding processes the full prompt during prefill, so TTFT scales with input length, from $0.139$s at $2$K to $20.3$s at $128$K.
MTP adds the MTP-head pass and is marginally slower than Regular at every length.
PIPO instead halves the effective prefill length by compressing every two input tokens into one, yielding a $1.65\times$ speedup over Regular at $2$K that grows to $2.64\times$ at $128$K ($20.3$s\,$\to$\,$7.69$s).
The relative gain increases with input length, because prefill cost dominates more strongly in long-context regimes, exactly where reasoning workloads live.

\paragraph{TPOT.}
Per-token cost is dominated by a single backbone forward pass; since MTP and PIPO both emit two tokens per pass, their TPOT is roughly half of Regular at all lengths.
Concretely, PIPO reaches a $2.07\times$ TPOT speedup at $2$K ($12.7$ vs.\ $26.3$) and $1.98\times$ at $128$K ($14.1$ vs.\ $27.9$), and is consistently the fastest because its compressed prefix also shrinks the KV cache.
Combined with the $2.64\times$ TTFT gain above, PIPO's speedups concentrate in the long-context regime, which dominates inference cost for modern reasoning models.

\subsection{Ablation Studies}\label{sec:experiments:ablation}
To understand the contribution of each PIPO component, we perform ablations on the architecture and training data.
Main results are shown in Table~\ref{tab:ablation}.
All ablations are run on Qwen3.5-4B.

\begin{table}[t]
\small
\centering
\caption{Ablations of the PIPO-SFT architecture and training data, evaluated by overall avg@4 / pass@4 on the four benchmarks of Table~\ref{tab:main}.}
\label{tab:ablation}
\begin{tabular}{l|cc}
\toprule
                   & Avg@4 & Pass@4 \\ \midrule
\rowcolor{blue!5}
PIPO-SFT (default)               & \textbf{45.28} & \textbf{65.01} \\
\;\;- linear compressor & 43.86  & 60.38  \\
\;\;- shortest response only & 42.88  & 59.87 \\
\;\;- random compressor init.   &  38.74 & 57.73 \\ \bottomrule
\end{tabular}
\end{table}

\paragraph{The compressor needs non-linearity.}
Replacing the MLP compressor with a single linear layer drops pass@4 by $4.63$ points ($65.01\!\to\!60.38$), confirming that a non-linear transformation is needed to fuse two heterogeneous token embeddings into one input latent that the backbone can consume.

\paragraph{Compressor initialization matters a lot.}
Initializing the MLP so that $f_\theta(a,b)\!\approx\!a\!+\!b$ keeps the backbone's input distribution close to its pretraining distribution at the start of training; random initialization causes the largest drop in the table ($-6.54$ avg@4, $-7.28$ pass@4), confirming our claim in Section~\ref{sec:method:pipo} that feeding out-of-distribution inputs to the backbone destabilizes early SFT.

\paragraph{SFT benefits from response diversity.}
Replacing our default ``all-correct responses'' data with the \emph{shortest} correct response per question drops pass@4 by $5.14$ points.
Keeping multiple correct trajectories per question therefore supplies useful diversity for the next-pair objective, even though they share the same final answer.

\begin{figure}[ht]
    \centering
    \includegraphics[width=0.8\linewidth]{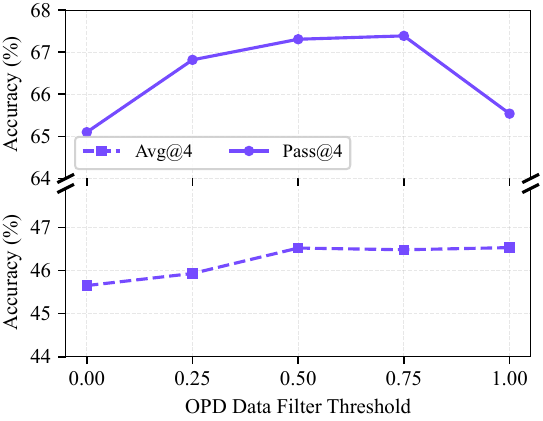}
    \caption{Effect of OPD data filtering by teacher correctness rate $\rho$ (the minimum fraction of correct teacher rollouts required to keep a question in the OPD set).}
    \label{fig:opd_data_ratio}
\end{figure}

\paragraph{OPD needs an accurate-but-not-trivial teacher.}
We keep only questions for which the teacher solves at least $\rho\!\in\!\{0\%, 25\%, 50\%, 75\%, 100\%\}$ of its four rollouts and re-train PIPO\,+\,OPD with each subset (Figure~\ref{fig:opd_data_ratio}).
Avg@4 grows monotonically with $\rho$, since a more correct teacher provides more reliable supervision for both the distillation loss and the confidence head.
However, pass@4 peaks at $\rho\!=\!50\%$ and then drops: aggressive filtering removes the hardest questions, leaving the student under-exposed to the cases that matter most for answer-space coverage.
We therefore use $\rho\!=\!50\%$ in Table~\ref{tab:main}: stable enough to learn from, while preserving difficult questions the student must eventually solve.

\begin{figure}[ht]
    \centering
    \includegraphics[width=0.8\linewidth]{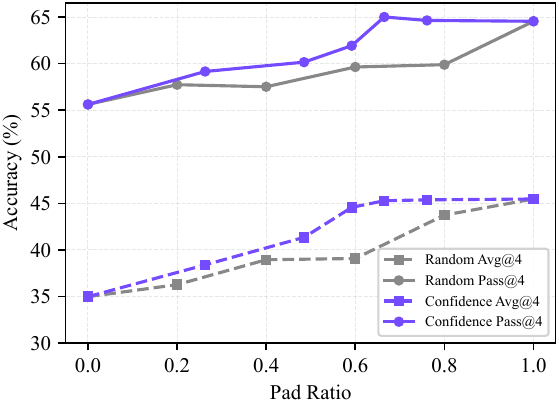}
    \caption{Overall avg@4 and pass@4 of PIPO-SFT on Qwen3.5-4B as a function of the \emph{pad ratio}, i.e., the fraction of draft positions whose input is replaced by the padding embedding at the next pair step.}
    \label{fig:acc_pad_ratio}
\end{figure}

\paragraph{The confidence head is non-trivial and recovers a sweet spot.}
We sweep the acceptance threshold $\tau_c\!\in\!\{0, 0.5, 0.8, 0.9, 0.95, 0.98, 1.0\}$ (yielding pad ratios from $0$ to $1$) and compare this \textit{Confidence} curve to a \textit{Random} baseline that matches the same average pad ratios via an unconditional coin flip (Figure~\ref{fig:acc_pad_ratio}).
Two observations stand out.

First, \textit{Confidence dominates Random at every intermediate pad ratio}: e.g., at pad $\sim\!0.6$ it reaches $61.93$ pass@4 vs.\ Random's $59.64$, and the same gap holds throughout.
This rules out the hypothesis that the head merely acts as an acceptance-rate knob; at the same acceptance budget it consistently rejects the drafts that matter and keeps the ones that help.

Second, \textit{Confidence peaks at an intermediate $\tau_c$, not at the conservative extreme}: pass@4 maxes out at $65.01$ for $\tau_c\!=\!0.95$ (pad $0.665$) and then drops to $64.55$ as $\tau_c\!\to\!1$ (single-token decoding).
Notably, this peak \emph{exceeds} the pad-$1$ baseline ($64.55$), meaning the accepted drafts under a well-tuned head do not merely preserve regular-decoding quality---they contribute additional, useful reasoning per step.
Avg@4, in contrast, grows monotonically with pad ratio, reflecting the usual precision--coverage trade-off; we use $\tau_c\!=\!0.95$ as the default in all main-table experiments.

%% file: sections/5-conclusion.tex
\section{Conclusion}\label{sec:conclusion}
We presented \textbf{PIPO}, a pair-in / pair-out framework for efficient LLM decoding that rests on two observations.
First, a latent compressor and an MTP head are mirror-image operations on the two sides of the backbone; combining them yields a symmetric pair-level interface, which halves the effective input length and doubles the per-step output.
Second, the on-policy distillation teacher plays the same role as the speculative-decoding verifier, so PIPO trains a lightweight confidence head with the teacher--student rejection-sampling ratio as a free label, amortizing the per-step verifier pass into a one-time training signal.
On four reasoning, coding, and long-context benchmarks, PIPO improves pass@4 over regular decoding by up to $+7.15$ pp on Qwen3.5-9B, with up to $2.64\times$ TTFT and $2.07\times$ TPOT speedups.

\section{Limitations}\label{sec:limitations}
PIPO has several limitations, which we leave to future work.
\textbf{First}, we study only the pair-in / pair-out setting. 
Larger compression factors may provide stronger speedups, but would be harder to model.
\textbf{Second}, our experiments are limited to 4B--9B models due to compute constraints. However, PIPO exhibits higher performance gains on the larger 9B backbone, suggesting that it may be even more effective for larger models.
\textbf{Third}, PIPO focuses on tasks with verifiable answers, and is not evaluated on open-ended generation tasks such as dialogue or creative writing.
\textbf{Fourth}, PIPO is studied only for text-only models. Extending latent compression to multi-modal settings may require modality-specific compressors.

%% file: sections/6-limitation_ethics.tex
\newpage
\section{Ethics Considerations}\label{sec:ethics}
PIPO is an inference-efficiency method for LLMs.
It does not introduce new training data sources or new user-facing capabilities by itself.
The main ethical impact is that faster decoding reduces inference cost and energy consumption, making reasoning models easier to deploy.
At the same time, improved efficiency may also lower the cost of harmful uses of LLMs.
The risks therefore largely follow those of the underlying base models and deployment settings.
We recommend using PIPO with the same safety filters, monitoring, and access controls applied to the original backbones.
Since confidence-based draft acceptance can affect generated content, practitioners should evaluate accuracy and safety on their target domains before deployment.

%% file: sections/9-appendix.tex
\section{Extended Related Work}\label{app:sec:relatedwork}

\subsection{LLM Reasoning}\label{app:sec:relatedwork:reasoning}
Chain-of-Thought (CoT) prompting~\cite{cot} encourages LLMs to produce explicit step-by-step traces before answering, and has been shown to substantially improve performance on complex tasks such as mathematics, code generation, long-form writing, and multimodal understanding~\cite{creval,revisor}.
Recent ``DeepThink'' models~\cite{o1,r1} go further by enclosing internal reasoning inside dedicated ``$\langle\mathtt{think}\rangle\langle\mathtt{/think}\rangle$'' tags before producing the final answer, a paradigm that has become the de-facto standard for the strongest open-weight reasoners.
Reinforcement-learning-based post-training further amplifies the reasoning capability of these models: group-based methods such as GRPO~\cite{grpo} and its successors DAPO~\cite{dapo} and GSPO~\cite{gspo} sample multiple candidate answers per prompt and re-weight them by correctness within each group.

\subsection{On-policy Distillation}\label{app:sec:relatedwork:opd}
On-policy distillation (OPD)~\cite{opd,rethinkingopd} is a hybrid of supervised fine-tuning and distillation in which the student rolls out responses under its own current policy and is then aligned, position-by-position, to a frozen teacher's distribution under a reverse-KL objective.
Compared with standard SFT, OPD removes the train--inference distribution shift; compared with full-trajectory RL, it uses a dense per-token signal and avoids costly reward modeling.
PIPO uses OPD both as a way to close the SFT--inference gap introduced by the pair-level interface and, more importantly, as the source of free supervision for its confidence head (Section~\ref{sec:method:opd}).

\subsection{Reasoning-efficient LLMs}\label{app:sec:relatedwork:efficient}
Long chain-of-thought reasoning inflates inference cost~\cite{cot,metacot}, and a growing literature studies how to shorten, compress, or adaptively terminate reasoning traces, ranging from prompting tricks~\cite{chainfodraft,sketchofthought} to architectural modifications and length-aware training~\cite{efficientsurvey,stopsurvey}.
PIPO is complementary to these efforts: it does not decide \emph{how much} reasoning the model should perform, but reduces the per-token cost of \emph{whatever} reasoning the model still produces.
The two directions can therefore be combined, e.g., a length-aware policy can be deployed on top of PIPO's pair-level decoder to compound the savings.

\section{Architecture Details}\label{app:sec:arch}

\subsection{Compressor Variants}\label{app:sec:arch:compressor}
The compressor maps a pair of consecutive token embeddings $(x^{2i}, x^{2i+1}) \in \mathbb{R}^{2H}$ into a single backbone-input latent $z^i \in \mathbb{R}^H$.
We support two drop-in variants with identical input/output signatures (Table~\ref{tab:compressor-variants}); the MLP variant is the default used in all main-table results.
Both are initialized so that $f_\theta([a;b]) \approx a + b$ at step $0$ (via zeroing the residual branch), which keeps the backbone's input distribution close to its pretraining distribution at the start of training (the ablation in Section~\ref{sec:experiments:ablation}).

\begin{table}[h]
\small
\centering
\caption{Compressor variants. $H$ is the backbone hidden size and $[\cdot\,;\cdot]$ denotes concatenation along the feature axis.}
\label{tab:compressor-variants}
\begin{tabular}{l|l}
\toprule
Variant & Formula \\ \midrule
Linear & $W [x^{2i}; x^{2i+1}] + b$ \\
MLP    & $W_2\, \mathrm{SiLU}(W_1 [x^{2i}; x^{2i+1}])$ \\ \bottomrule
\end{tabular}
\end{table}

\subsection{MTP Head}\label{app:sec:arch:mtp}
The MTP head is a single full-attention decoder layer attached \emph{after} the last backbone layer.
Concretely, given the backbone hidden state $h_b^i$ and the embedding of the just-decoded backbone token $x^{2i+2}$, we apply
$h_d^i = \mathrm{Layer}\bigl(W_{\mathrm{fc}}\,[\,\mathrm{RMSNorm}(h_b^i)\,;\,\mathrm{RMSNorm}(x^{2i+2})\,]\bigr)$,
where $\mathrm{Layer}$ is a Qwen3.5 decoder block sharing the backbone's hyperparameters and $W_{\mathrm{fc}}\!:\!\mathbb{R}^{2H}\!\to\!\mathbb{R}^{H}$ is a learnable projection.
The same frozen LM head is then applied to $h_d^i$ to produce the draft-token distribution $p_d^{2i+3}$.
For the off-the-shelf MTP-equipped backbones used in this paper, the MTP layer ships pre-trained from Qwen3.5; PIPO only LoRA-adapts its attention and MLP projections and fully trains the small norm/projection modules (Section~\ref{app:sec:training:sft}).

\subsection{Confidence Head}\label{app:sec:arch:conf}
The confidence head $g_\phi$ takes the concatenated pair of backbone and MTP hidden states at pair step $i$ and returns a scalar acceptance probability for the draft token:
\begin{equation*}
    c^i = \sigma\!\left( W_2\,\mathrm{SiLU}\bigl(W_1\,\mathrm{RMSNorm}([h_b^i; h_d^i])\bigr)\right),
\end{equation*}
with $W_1\!:\!\mathbb{R}^{2H}\!\to\!\mathbb{R}^{H}$ and $W_2\!:\!\mathbb{R}^{H}\!\to\!\mathbb{R}$ (no bias).
On a $4$B backbone this adds $\sim\!6.6$M parameters ($\approx 0.16\%$ of the model), and a single forward pass through the head is orders of magnitude cheaper than a full backbone verifier pass.

\section{Training Implementation Details}\label{app:sec:training}

\subsection{SFT Data}\label{app:sec:training:data}
\begin{figure}[h]
\centering
\includegraphics[width=0.8\linewidth]{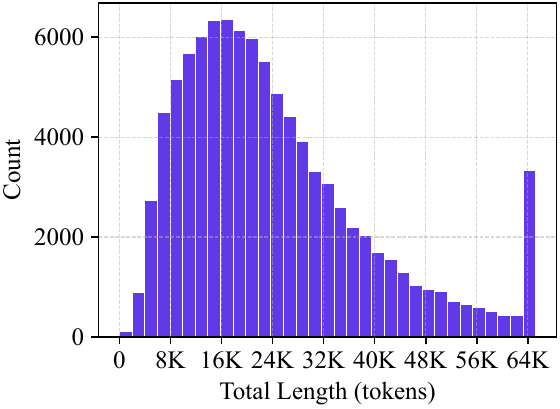}
\caption{Token length distribution of the SFT corpus ($95{,}969$ Qwen3.5-9B trajectories on DAPO-Math and Codeforces).}
\label{app:fig:sft_data}
\end{figure}

SFT trajectories are obtained by sampling four responses per question from the Qwen3.5-9B teacher on the union of DAPO-Math and Codeforces and keeping all correct trajectories, yielding $95{,}969$ samples in total.
Figure~\ref{app:fig:sft_data} shows the length distribution: mean $24.9$K, median $21.6$K, standard deviation $14.9$K tokens, with a hard cap of $64$K imposed at tokenization.
The long right tail is the main reason why we cap evaluation at the $32$K-slot budget shared with all baselines.

\subsection{SFT Training}\label{app:sec:training:sft}
We use ms-swift~\cite{msswift} with LoRA~\cite{lora} (rank $64$, $\alpha\!=\!128$, dropout $0.05$) on the backbone projections $\{q,k,v,o,\mathrm{gate},\mathrm{up},\mathrm{down}\}$ and, by name-suffix match, the same projections inside the MTP decoder layer.
Beyond the LoRA adapters, we fully train the compressor MLP, the MTP projection $W_{\mathrm{fc}}$, the MTP pre/post norms, and the confidence head; the LM head is tied to the input embedding and frozen.
We train for $2$ epochs on $8$ H20-141G GPUs (per-device batch size $1$, gradient accumulation $16$) with DeepSpeed ZeRO-2, AdamW at learning rate $1\!\times\!10^{-4}$ ($5\%$ linear warmup, cosine annealing), max sequence length $64$K, and flash-attention~2.
The MTP-loss weight $\lambda_{\mathrm{mtp}}$ and the confidence BCE weight $\lambda_{\mathrm{conf}}$ are both fixed at $1$.

\paragraph{Random PAD injection.}
To expose the model to rejected-draft pairs at training time, we randomly inject the padding embedding at draft positions: at every step we sample $\rho\!\sim\!\mathrm{Uniform}(0, \rho_{\max})$ with $\rho_{\max}\!=\!0.25$, then independently mark a $\rho$ fraction of eligible pairs $(x^{2p}, x^{2p+1})$ in the response region for splitting into $\{(x^{2p}, x^{\mathrm{pad}}),\,(x^{2p+1}, x^{\mathrm{pad}})\}$.
Labels at PAD positions are masked from the cross-entropy and confidence losses.
We additionally guarantee that PAD tokens always land at the odd position of a pair (so prompt/response boundaries align with pair boundaries) and that the total sequence length stays even, with the attention mask updated accordingly.

\paragraph{Confidence head target (SFT).}
During SFT, the teacher's distribution at every label position is a one-hot $\delta_{y_t}$ over the ground-truth token, so the rejection-sampling acceptance probability collapses to
\begin{align*}
    \sum_{y} \min\bigl(p_s(y),\,p_t(y)\bigr)
        &= \min\bigl(p_s(y_t),\,1\bigr) \\
        &= p_s(y_t),
\end{align*}
i.e., the student's own probability on the gold draft token.
The SFT-stage BCE target for the confidence head is therefore $p_d^{2i+3}(x^{2i+3})$, which is exactly the form used in Equation~(7) of the main paper.
Crucially, this target collapses to the same quantity as the OPD-stage rejection-sampling acceptance target in the deterministic-teacher limit, so the head transfers from SFT to OPD without a parameter reset.

\subsection{OPD Training}\label{app:sec:training:opd}
OPD is a three-stage pipeline per micro-batch.
\textbf{(1) Rollout}: each question is rolled out by the SFT student under its own pair-level decoder via an SGLang colocate engine~\cite{sglang}, with the radix cache disabled to preserve PAD-augmentation determinism.
\textbf{(2) Teacher forward (PAD compaction)}: because the uncompressed Qwen3.5-9B teacher has never seen mid-sequence PAD tokens, we strip every PAD from the rolled-out trajectory, run the teacher on the clean sequence, and then re-map the teacher's per-position log-probabilities back to the original (PAD-augmented) positions via a cumulative-sum lookup, where each PAD position inherits the teacher's distribution conditioned on the immediately preceding non-PAD prefix, which is the correct conditional under our pair-in semantics.
\textbf{(3) Student forward and loss}: the student is re-forwarded on the same trajectory in compressed (pair-level) mode so that every trainable module (compressor, MTP head, LoRA on backbone and MTP, confidence head) receives gradient.

The default loss is a Monte-Carlo reverse-KL on the on-policy sampled tokens,
\begin{equation*}
    \mathcal{L}_{\mathrm{distill}} = \mathbb{E}_{y\sim p_s}\!\left[\log p_s(y) - \log p_t(y)\right],
\end{equation*}
applied separately at even positions (backbone head) and odd positions (MTP head).
We additionally support a top-$k$ union mode (which restricts the divergence to the union of the student top-$k$, teacher top-$k$, and the sampled label, with $k\!=\!32$) and a full-vocab JSD mode; these are used only in early ablations.
All KL/JSD computations are chunked along the token dimension with chunk size $2048$, capping peak memory at $O(\mathrm{chunk}\,\times V)$ instead of $O(T \times V)$.

\paragraph{Confidence head target (OPD).}
At the rolled-out draft position $y\!=\!x^{2i+3}$, the per-token rejection-sampling acceptance probability is
\begin{align*}
    \alpha_i &= \min\!\left(\frac{p_t(y)}{p_s(y)},\,1\right) \\
             &= \exp\Bigl(-\bigl[\log p_s(y) - \log p_t(y)\bigr]_+\Bigr),
\end{align*}
which is exactly the per-position quantity already computed by the sampled-KL loss above (modulo a clamp and an $\exp$).
The confidence head is therefore trained with a BCE loss against $\alpha_i$ as a one-time signal that recycles the same teacher/student forward passes used by the distillation loss; no extra teacher or student calls are introduced.
We detach $\alpha_i$ from the autograd graph before BCE so that the head cannot leak gradient into the student's logits via its own target.

\paragraph{OPD hyperparameters.}
We use $1$ epoch, per-device batch size $1$ with gradient accumulation $4$, the same optimizer schedule as SFT, and LoRA rank $64$ ($\alpha\!=\!128$) on the same modules.
Training runs on $8$ H20-141G GPUs with $\mathrm{tp}\!=\!1$ and SGLang's colocate rollout (model weights kept GPU-resident across rollout/training switches).
We disable cross-sample padding inside the OPD chunk loop by running each sample through its own $B\!=\!1$ chunk and averaging.

\section{Evaluation Implementation Details}\label{app:sec:evaluation}

\subsection{Inference Setup}\label{app:sec:evaluation:setup}
All decoding runs use SGLang~\cite{sglang} (with our LatentMTP extensions for PIPO) on $8\times$ NVIDIA H20-141G GPUs, with data parallelism size $8$, tensor parallelism size $1$, and at most $64$ concurrent requests per GPU.
The response budget is set to $32$K slots (Section~\ref{sec:experiments:efficiency}) for every method, and the sampling parameters are fixed across methods as reported in Section~\ref{sec:experiments}.

\subsection{Baseline Configurations}\label{app:sec:evaluation:baselines}
\paragraph{Regular.}
Plain autoregressive decoding through the Qwen3.5 SGLang backend.

\paragraph{MTP without verification.}
We run the Qwen3.5 model on the same SGLang backend with PIPO's pair-out path enabled (so Qwen3.5's MTP head produces a draft per backbone step), but without any verification; every draft token is therefore committed unconditionally.

\paragraph{EAGLE-2.}
We enable SGLang's tree-based speculative decoding (\texttt{NEXTN} algorithm) with $3$ speculative steps, $\mathrm{top\text{-}k}\!=\!1$ at every draft level, and $4$ draft tokens per verifier call.
The draft head is the off-the-shelf Qwen3.5 MTP head; the verifier is the full backbone.

\paragraph{PIPO.}
Both PIPO-SFT and PIPO\,+\,OPD are deployed with the confidence head active and $\tau_c\!=\!0.95$, selected as the pass@4 sweet spot in the confidence-threshold sweep of Section~\ref{sec:experiments:ablation}.

\subsection{Answer Evaluators}\label{app:sec:evaluation:answer}
Models are instructed to put their final answer inside ``$\backslash\mathtt{boxed}\{\cdot\}$''.
We extract the answer with a regular expression and, when several boxed expressions are present, take the last one.
Mathematical answers (AIME 2025) are compared to the ground truth with the math-verify library;\footnote{\url{https://github.com/huggingface/Math-Verify}} multiple-choice answers (GPQA-Diamond, LongBench v2) use exact string match against the gold option label.
LiveCodeBench v6 is execution-based: the extracted code is compiled and run against the benchmark's hidden test suite, and the problem is counted as correct only if all tests pass.

\section{Additional Analyses}\label{app:sec:experiments}
We complement the quantitative results of Section~\ref{sec:experiments} with a probe of two of PIPO's most distinctive components, the pair-in compressor (Appendix~\ref{app:sec:experiments:compressor}) and the padding token (Appendix~\ref{app:sec:experiments:pad}).
All analyses use the PIPO\,+\,OPD checkpoint on Qwen3.5-4B, pooled over $12$ prompts ($4$ each from AIME 2025, GPQA-Diamond, and LiveCodeBench v6), for a total of $1{,}372$ token pairs.

\subsection{What does the compressor learn?}\label{app:sec:experiments:compressor}
The compressor $f_\theta\colon \mathbb{R}^{2H}\!\to\!\mathbb{R}^{H}$ is initialized so that $f_\theta([a;b]) \approx a+b$ (Appendix~\ref{app:sec:arch:compressor}); a natural question is what it deviates to after training.
We answer this with three complementary probes: \emph{(i)} per-input Jacobian-norm sensitivity, \emph{(ii)} swap-test cosine, and \emph{(iii)} compressor-vs-sum alignment.

\begin{figure}[h]
\centering
\includegraphics[width=0.8\linewidth]{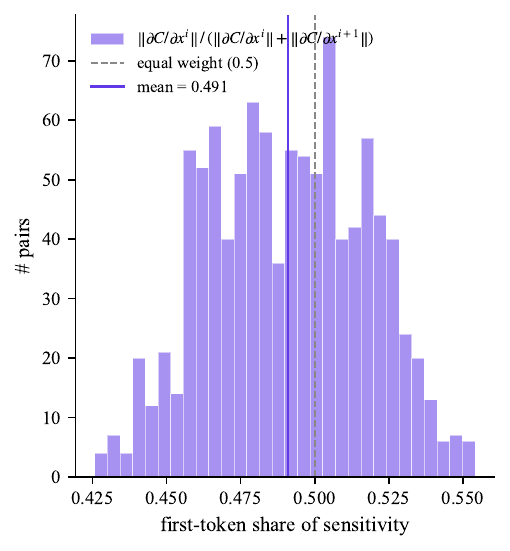}
\caption{Per-pair sensitivity share of the first input, $\|\partial f_\theta/\partial x^{2i}\|/(\|\partial f_\theta/\partial x^{2i}\|+\|\partial f_\theta/\partial x^{2i+1}\|)$. The distribution is tightly concentrated near $0.5$ (mean $0.491$): the compressor weights both pair positions roughly equally.}
\label{app:fig:compressor_sensitivity}
\end{figure}

\paragraph{The compressor is position-unbiased.}
For each pair we measure the share of the output sensitivity attributable to the first input, ${\|\partial f_\theta/\partial x^{2i}\|}\,/\,({\|\partial f_\theta/\partial x^{2i}\|}+{\|\partial f_\theta/\partial x^{2i+1}\|})$.
Figure~\ref{app:fig:compressor_sensitivity} shows that this ratio concentrates tightly around $0.49$, statistically indistinguishable from the symmetric value $0.5$.
The compressor therefore systematically over-weights neither the leading nor the trailing token in a pair, mirroring its symmetric additive initialization rather than collapsing onto one position.

\begin{figure}[h]
\centering
\includegraphics[width=0.8\linewidth]{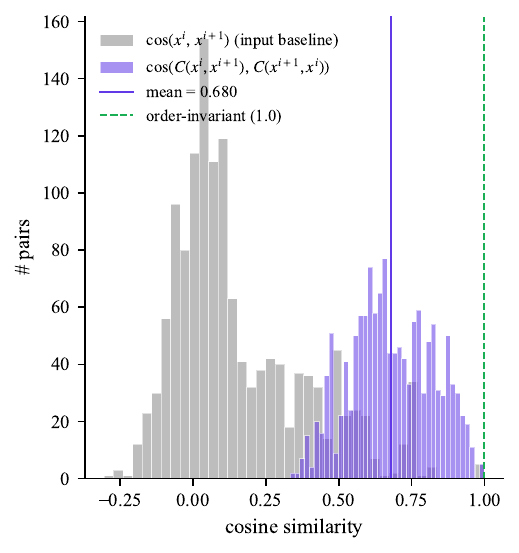}
\caption{Swap-test cosine $\cos(f_\theta([a;b]),\,f_\theta([b;a]))$ (purple, mean $0.68$) against the input baseline $\cos(a, b)$ (gray, mean $0.16$). A fully order-invariant compressor would sit at $1.0$ (green dashed); the trained compressor sits much lower, showing that it encodes slot identity.}
\label{app:fig:compressor_position_sensitivity}
\end{figure}

\paragraph{The compressor is position-aware.}
If the compressor were a symmetric function (e.g., its sum-initialized starting point), swapping the two inputs would leave the output unchanged.
We measure $\cos\bigl(f_\theta([x^{2i};x^{2i+1}]),\,f_\theta([x^{2i+1};x^{2i}])\bigr)$ for every pair and compare to the input baseline $\cos(x^{2i}, x^{2i+1})$ that describes how similar the two raw embeddings already are.
Figure~\ref{app:fig:compressor_position_sensitivity} shows the swap cosine concentrates at $0.68$, well below the order-invariant value $1.0$ and far above the input baseline of $0.16$.
The compressor therefore moves substantially away from the symmetric solution during training and learns to encode \emph{which} slot each token occupies in addition to \emph{which} two tokens it received.

\begin{figure}[h]
\centering
\includegraphics[width=0.8\linewidth]{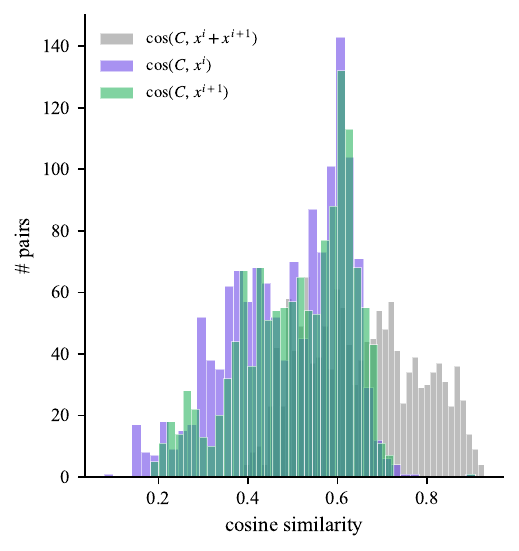}
\caption{Cosine similarity between the compressor output and three references: the naive sum $x^{2i}+x^{2i+1}$ (gray), the first input alone (purple), and the second input alone (green). The output aligns most with the sum but stays clearly below $1.0$.}
\label{app:fig:compressor_cos_distribution}
\end{figure}

\paragraph{The compressor learns beyond a sum.}
The previous two probes still leave open whether the trained compressor is essentially an additive map with a learned positional twist.
Figure~\ref{app:fig:compressor_cos_distribution} shows that the compressor output is closer to the additive baseline $x^{2i}+x^{2i+1}$ (mean cosine $0.65$) than to either constituent alone ($0.49$ and $0.51$), but the gap from $1.0$ is large and the output magnitude is sharply attenuated ($\|f_\theta(\cdot)\|/\|x^{2i}+x^{2i+1}\| \approx 0.50$ on average).
Taken together with the swap-test, this confirms that $f_\theta$ has moved well past its $a+b$ initialization---it preserves the additive geometry that keeps the backbone in distribution while learning a sharper, position-aware projection that brings the pair embedding into a regime the backbone can decode at the pair granularity.

\subsection{What is the role of the padding token?}\label{app:sec:experiments:pad}
The padding embedding plays a structural role at both training time (random PAD injection, Appendix~\ref{app:sec:training:sft}) and inference time (the next input pair after a rejected draft is $(x^{2i+2}, x^{\mathrm{pad}})$, Section~\ref{sec:method:pipo}).
For PIPO to keep the same pair-level interface across both regimes, the compressor should treat a PAD-padded pair as ``the surviving token alone'' rather than as an out-of-distribution input.
We verify this from two angles.

\begin{figure}[h]
\centering
\includegraphics[width=0.8\linewidth]{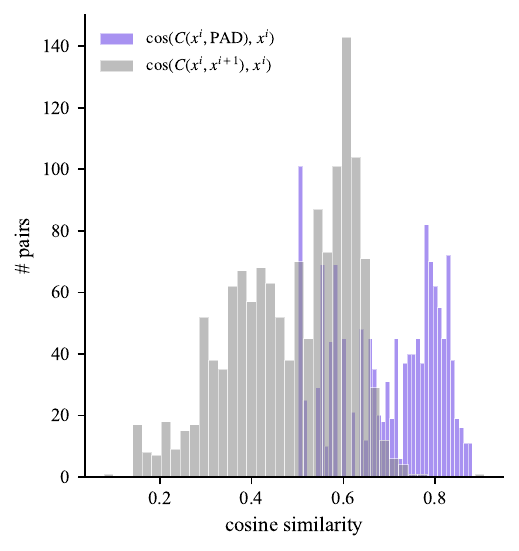}
\caption{Effect of replacing the second pair token with PAD. Purple: $\cos(f_\theta([x^{2i};\mathrm{PAD}]),\,x^{2i})$, mean $0.70$. Gray: $\cos(f_\theta([x^{2i};x^{2i+1}]),\,x^{2i})$, mean $0.49$. A PAD-injected pair aligns much more strongly with the surviving token, i.e., PAD is effectively ignored.}
\label{app:fig:compressor_pad_role}
\end{figure}

\paragraph{PAD is effectively ignored.}
Figure~\ref{app:fig:compressor_pad_role} replaces the second token of every pair by the PAD embedding and compares the compressor output to the surviving first token.
The PAD-injected output aligns much more strongly with the surviving token ($\cos = 0.70$) than the unmodified output does ($\cos = 0.49$): when one slot is PAD, the compressor effectively delegates the pair latent to the non-PAD slot.
The same behavior holds in reverse---putting PAD in the first slot raises $\cos(f_\theta([\mathrm{PAD};b]), b)$ to $0.60$, well above the additive baseline.

\begin{figure}[h]
\centering
\includegraphics[width=0.8\linewidth]{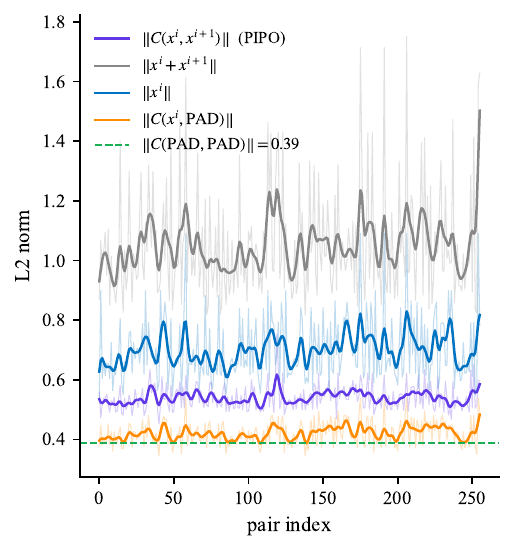}
\caption{Compressor output norm along the longest probe trajectory, with all-PAD norm overlaid as a green dashed line. The all-PAD output is an order of magnitude smaller than every alternative, making it act as a near-zero signal to the backbone.}
\label{app:fig:compressor_norms_longest}
\end{figure}

\paragraph{All-PAD pairs collapse to a near-zero signal.}
Figure~\ref{app:fig:compressor_norms_longest} plots the magnitude of the compressor output along a representative trajectory and compares it to several controls.
The pure-PAD output $\|f_\theta([\mathrm{PAD};\mathrm{PAD}])\| = 0.39$ (green dashed) is an order of magnitude below every other curve, including the PAD-on-one-side output (orange).
This makes the all-PAD pair behave almost like a no-op KV-cache entry for the backbone: it has the right shape but carries vanishing signal, which is exactly the property needed for PIPO to fall back to single-token decoding under aggressive rejection (Section~\ref{sec:experiments:ablation}) without polluting the hidden state of subsequent slots.

%% file: custom.bib
@article{colar,
  title={Think silently, think fast: Dynamic latent compression of llm reasoning chains},
  author={Tan, Wenhui and Li, Jiaze and Ju, Jianzhong and Luo, Zhenbo and Song, Ruihua and Luan, Jian},
  journal={Advances in Neural Information Processing Systems},
  volume={38},
  pages={4646--4668},
  year={2026}
}

@misc{aime,
  author       = {AMC},
  howpublished = {\url{https://artofproblemsolving.com/wiki/index.php/American_Invitational_Mathematics_Examination}},
  year         = {2025},
}

@article{chainfodraft,
  title={Chain of draft: Thinking faster by writing less},
  author={Xu, Silei and Xie, Wenhao and Zhao, Lingxiao and He, Pengcheng},
  journal={arXiv preprint arXiv:2502.18600},
  year={2025}
}

@article{coconut,
  title={Training large language models to reason in a continuous latent space},
  author={Hao, Shibo and Sukhbaatar, Sainbayar and Su, DiJia and Li, Xian and Hu, Zhiting and Weston, Jason and Tian, Yuandong},
  journal={arXiv preprint arXiv:2412.06769},
  year={2024}
}

@article{cot,
  title={Chain-of-thought prompting elicits reasoning in large language models},
  author={Wei, Jason and Wang, Xuezhi and Schuurmans, Dale and Bosma, Maarten and Xia, Fei and Chi, Ed and Le, Quoc V and Zhou, Denny and others},
  journal={Advances in neural information processing systems},
  volume={35},
  pages={24824--24837},
  year={2022}
}

@article{creval,
  title={Evaluating Text Creativity across Diverse Domains: A Dataset and Large Language Model Evaluator},
  author={Cao, Qian and Wang, Xiting and Yuan, Yuzhuo and Liu, Yahui and Luo, Fang and Song, Ruihua},
  journal={arXiv preprint arXiv:2505.19236},
  year={2025}
}

@article{dapo,
  title={Dapo: An open-source llm reinforcement learning system at scale},
  author={Yu, Qiying and Zhang, Zheng and Zhu, Ruofei and Yuan, Yufeng and Zuo, Xiaochen and Yue, Yu and Fan, Tiantian and Liu, Gaohong and Liu, Lingjun and Liu, Xin and others},
  journal={arXiv preprint arXiv:2503.14476},
  year={2025}
}

@article{deepseekv3,
  title={Deepseek-v3 technical report},
  author={Liu, Aixin and Feng, Bei and Xue, Bing and Wang, Bingxuan and Wu, Bochao and Lu, Chengda and Zhao, Chenggang and Deng, Chengqi and Zhang, Chenyu and Ruan, Chong and others},
  journal={arXiv preprint arXiv:2412.19437},
  year={2024}
}

@article{efficientsurvey,
    title={Efficient Reasoning Models: A Survey},
    author={Feng, Sicheng and Fang, Gongfan and Ma, Xinyin and Wang, Xinchao},
    journal={arXiv preprint arXiv:2504.10903},
    year={2025},
}

@inproceedings{eagle, 
	author = {Yuhui Li and Fangyun Wei and Chao Zhang and Hongyang Zhang}, 
	title = {{EAGLE}: Speculative Sampling Requires Rethinking Feature Uncertainty}, 
	booktitle = {International Conference on Machine Learning},
	year = {2024}
}

@inproceedings{eagle2,
  title={Eagle-2: Faster inference of language models with dynamic draft trees},
  author={Li, Yuhui and Wei, Fangyun and Zhang, Chao and Zhang, Hongyang},
  booktitle={Proceedings of the 2024 conference on empirical methods in natural language processing},
  pages={7421--7432},
  year={2024}
}

@inproceedings{eagle3,
    author = {Yuhui Li and Fangyun Wei and Chao Zhang and Hongyang Zhang},
    title = {{EAGLE-3}: Scaling up Inference Acceleration of Large Language Models via Training-Time Test}, 
    booktitle = {Annual Conference on Neural Information Processing Systems},
    year = {2025}
}

@inproceedings{gpqa,
title={GPQA: A Graduate-Level Google-Proof Q\&A Benchmark},
author={David Rein and Betty Li Hou and Asa Cooper Stickland and Jackson Petty and Richard Yuanzhe Pang and Julien Dirani and Julian Michael and Samuel R. Bowman},
booktitle={First Conference on Language Modeling},
year={2024},
url={https://openreview.net/forum?id=Ti67584b98}
}

@article{grpo,
  title={Deepseekmath: Pushing the limits of mathematical reasoning in open language models},
  author={Shao, Zhihong and Wang, Peiyi and Zhu, Qihao and Xu, Runxin and Song, Junxiao and Bi, Xiao and Zhang, Haowei and Zhang, Mingchuan and Li, YK and Wu, Y and others},
  journal={arXiv preprint arXiv:2402.03300},
  year={2024}
}

@article{gspo,
  title={Group sequence policy optimization},
  author={Zheng, Chujie and Liu, Shixuan and Li, Mingze and Chen, Xiong-Hui and Yu, Bowen and Gao, Chang and Dang, Kai and Liu, Yuqiong and Men, Rui and Yang, An and others},
  journal={arXiv preprint arXiv:2507.18071},
  year={2025}
}

@article{latentspacesurvey,
  title={The latent space: Foundation, evolution, mechanism, ability, and outlook},
  author={Yu, Xinlei and Chen, Zhangquan and He, Yongbo and Fu, Tianyu and Yang, Cheng and Xu, Chengming and Ma, Yue and Hu, Xiaobin and Cao, Zhe and Xu, Jie and others},
  journal={arXiv preprint arXiv:2604.02029},
  year={2026}
}

@article{livecodebench,
  title={Livecodebench: Holistic and contamination free evaluation of large language models for code},
  author={Jain, Naman and Han, King and Gu, Alex and Li, Wen-Ding and Yan, Fanjia and Zhang, Tianjun and Wang, Sida and Solar-Lezama, Armando and Sen, Koushik and Stoica, Ion},
  journal={arXiv preprint arXiv:2403.07974},
  year={2024}
}

@inproceedings{longbenchv2,
  title={Longbench v2: Towards deeper understanding and reasoning on realistic long-context multitasks},
  author={Bai, Yushi and Tu, Shangqing and Zhang, Jiajie and Peng, Hao and Wang, Xiaozhi and Lv, Xin and Cao, Shulin and Xu, Jiazheng and Hou, Lei and Dong, Yuxiao and others},
  booktitle={Proceedings of the 63rd Annual Meeting of the Association for Computational Linguistics (Volume 1: Long Papers)},
  pages={3639--3664},
  year={2025}
}

@article{lora,
  title={Lora: Low-rank adaptation of large language models.},
  author={Hu, Edward J and Shen, Yelong and Wallis, Phillip and Allen-Zhu, Zeyuan and Li, Yuanzhi and Wang, Shean and Wang, Lu and Chen, Weizhu and others},
  journal={ICLR},
  volume={1},
  number={2},
  pages={3},
  year={2022}
}

@article{medusa,
  title={Medusa: Simple llm inference acceleration framework with multiple decoding heads},
  author={Cai, Tianle and Li, Yuhong and Geng, Zhengyang and Peng, Hongwu and Lee, Jason D and Chen, Deming and Dao, Tri},
  journal={arXiv preprint arXiv:2401.10774},
  year={2024}
}

@article{metacot,
  title={Towards System 2 Reasoning in LLMs: Learning How to Think With Meta Chain-of-Though},
  author={Xiang, Violet and Snell, Charlie and Gandhi, Kanishk and Albalak, Alon and Singh, Anikait and Blagden, Chase and Phung, Duy and Rafailov, Rafael and Lile, Nathan and Mahan, Dakota and others},
  journal={arXiv preprint arXiv:2501.04682},
  year={2025}
}

@article{mimov2flash,
  title={MiMo-V2-Flash Technical Report},
  author={Xiao, Bangjun and Xia, Bingquan and Yang, Bo and Gao, Bofei and Shen, Bowen and Zhang, Chen and He, Chenhong and Lou, Chiheng and Luo, Fuli and Wang, Gang and others},
  journal={arXiv preprint arXiv:2601.02780},
  year={2026}
}

@misc{msswift,
  title={SWIFT:A Scalable lightWeight Infrastructure for Fine-Tuning},
  author={Yuze Zhao and Jintao Huang and Jinghan Hu and Xingjun Wang and Yunlin Mao and Daoze Zhang and Zeyinzi Jiang and Zhikai Wu and Baole Ai and Ang Wang and Wenmeng Zhou and Yingda Chen},
  year={2024},
  eprint={2408.05517},
  archivePrefix={arXiv},
  primaryClass={cs.CL},
  url={https://arxiv.org/abs/2408.05517},
}

@article{multiplexthinking,
  title   = {Multiplex Thinking: Reasoning via Token-wise Branch-and-Merge},
  author  = {Tang, Yao and Dong, Li and Hao, Yaru and Dong, Qingxiu and Wei, Furu and Gu, Jiatao},
  journal = {arXiv preprint arXiv:2601.08808},
  year    = {2026}
}

@article{o1,
  title={Openai o1 system card},
  author={Jaech, Aaron and Kalai, Adam and Lerer, Adam and Richardson, Adam and El-Kishky, Ahmed and Low, Aiden and Helyar, Alec and Madry, Aleksander and Beutel, Alex and Carney, Alex and others},
  journal={arXiv preprint arXiv:2412.16720},
  year={2024}
}

@inproceedings{opd,
  title={On-policy distillation of language models: Learning from self-generated mistakes},
  author={Agarwal, Rishabh and Vieillard, Nino and Zhou, Yongchao and Stanczyk, Piotr and Ramos Garea, Sabela and Geist, Matthieu and Bachem, Olivier},
  booktitle={International Conference on Learning Representations},
  volume={2024},
  pages={21246--21263},
  year={2024}
}

@article{qwen3,
  title={Qwen3 technical report},
  author={Yang, An and Li, Anfeng and Yang, Baosong and Zhang, Beichen and Hui, Binyuan and Zheng, Bo and Yu, Bowen and Gao, Chang and Huang, Chengen and Lv, Chenxu and others},
  journal={arXiv preprint arXiv:2505.09388},
  year={2025}
}

@article{r1,
  title={Deepseek-r1: Incentivizing reasoning capability in llms via reinforcement learning},
  author={Guo, Daya and Yang, Dejian and Zhang, Haowei and Song, Junxiao and Zhang, Ruoyu and Xu, Runxin and Zhu, Qihao and Ma, Shirong and Wang, Peiyi and Bi, Xiao and others},
  journal={arXiv preprint arXiv:2501.12948},
  year={2025}
}

@article{rethinkingopd,
  title={Rethinking on-policy distillation of large language models: Phenomenology, mechanism, and recipe},
  author={Li, Yaxuan and Zuo, Yuxin and He, Bingxiang and Zhang, Jinqian and Xiao, Chaojun and Qian, Cheng and Yu, Tianyu and Gao, Huan-ang and Yang, Wenkai and Liu, Zhiyuan and others},
  journal={arXiv preprint arXiv:2604.13016},
  year={2026}
}

@article{revisor,
  title={REVISOR: Beyond Textual Reflection, Towards Multimodal Introspective Reasoning in Long-Form Video Understanding},
  author={Li, Jiaze and Yin, Hao and Tan, Wenhui and Chen, Jingyang and Xu, Boshen and Qu, Yuxun and Chen, Yijing and Ju, Jianzhong and Luo, Zhenbo and Luan, Jian},
  journal={arXiv preprint arXiv:2511.13026},
  year={2025}
}

@article{rmsnorm,
  title={Root mean square layer normalization},
  author={Zhang, Biao and Sennrich, Rico},
  journal={Advances in neural information processing systems},
  volume={32},
  year={2019}
}

@article{sglang,
  title={Sglang: Efficient execution of structured language model programs},
  author={Zheng, Lianmin and Yin, Liangsheng and Xie, Zhiqiang and Sun, Chuyue Livia and Huang, Jeff and Yu, Cody Hao and Cao, Shiyi and Kozyrakis, Christos and Stoica, Ion and Gonzalez, Joseph E and others},
  journal={Advances in neural information processing systems},
  volume={37},
  pages={62557--62583},
  year={2024}
}

@article{singlethreaded,
  title={Llms are single-threaded reasoners: Demystifying the working mechanism of soft thinking},
  author={Wu, Junhong and Lu, Jinliang and Ren, Zixuan and Hu, Gangqiang and Wu, Zhi and Dai, Dai and Wu, Hua},
  journal={arXiv preprint arXiv:2508.03440},
  year={2025}
}

@article{sketchofthought,
  title={Sketch-of-thought: Efficient llm reasoning with adaptive cognitive-inspired sketching},
  author={Aytes, Simon A and Baek, Jinheon and Hwang, Sung Ju},
  journal={arXiv preprint arXiv:2503.05179},
  year={2025}
}

@article{softthinking,
  title={Soft thinking: Unlocking the reasoning potential of llms in continuous concept space},
  author={Zhang, Zhen and He, Xuehai and Yan, Weixiang and Shen, Ao and Zhao, Chenyang and Wang, Shuohang and Shen, Yelong and Wang, Xin Eric},
  journal={arXiv preprint arXiv:2505.15778},
  year={2025}
}

@inproceedings{speculativedecoding,
  title={Fast inference from transformers via speculative decoding},
  author={Leviathan, Yaniv and Kalman, Matan and Matias, Yossi},
  booktitle={International Conference on Machine Learning},
  pages={19274--19286},
  year={2023},
  organization={PMLR}
}

@article{stopsurvey,
  title={Stop overthinking: A survey on efficient reasoning for large language models},
  author={Sui, Yang and Chuang, Yu-Neng and Wang, Guanchu and Zhang, Jiamu and Zhang, Tianyi and Yuan, Jiayi and Liu, Hongyi and Wen, Andrew and Zhong, Shaochen and Chen, Hanjie and others},
  journal={arXiv preprint arXiv:2503.16419},
  year={2025}
}

@misc{codeforces,
      title={CodeForces}, 
      author={Guilherme Penedo and Anton Lozhkov and Hynek Kydlíček and Loubna Ben Allal and Edward Beeching and Agustín Piqueres Lajarín and Quentin Gallouédec and Nathan Habib and Lewis Tunstall and Leandro von Werra},
      year={2025},
      publisher = {Hugging Face},
      journal = {Hugging Face repository},
      howpublished = {\url{https://huggingface.co/datasets/open-r1/codeforces}}
}

@misc{qwen35,
    title = {Qwen3.5: Accelerating Productivity with Native Multimodal Agents},
    url = {https://qwen.ai/blog?id=qwen3.5},
    author = {Qwen Team},
    month = {February},
    year = {2026}
}
